\RequirePackage{fix-cm}
\documentclass[twocolumn]{svjour3}          
\usepackage{graphicx}
\usepackage{array}
\newcolumntype{C}[1]{>{\centering\let\newline\\\arraybackslash\hspace{0pt}}m{#1}}
\usepackage[justification=centering]{caption}

 \usepackage{mathptmx}      
\usepackage{amssymb, amsmath}
\usepackage{graphicx} 
\usepackage{algorithm}
\usepackage{algorithmic}
\usepackage{hyperref}
\usepackage{caption}
\usepackage{enumerate}
\usepackage{multirow}
\usepackage{booktabs}
\usepackage{stfloats}
\usepackage{subcaption}
\usepackage{sidecap}

\usepackage[numbers]{natbib}
\usepackage{mathpazo}
\smartqed  
\usepackage{graphicx}
\begin{document}

\title{Active Mining of Parallel Video Streams}
\author{Samaneh~Khoshrou 
      \and  Jaime~S.~Cardoso     \and Lu\'{i}s~F.~Teixeira 
}


\institute{ S. Khoshrou, J. S. Cardoso \at INESC TEC (formerly INESC Porto)
         \at
              Rua Doutor Roberto Frias 378, 4200-465 Oporto \\
              Tel.: +351-222094000 \\
                           \email{samaneh.khoshrou@inescporto.pt\\
                           jaime.cardoso@inescporto.pt}           
        \and  L. F.~Teixeira\\ 
         Faculdade de Engenharia da Universidade do Porto (FEUP)\\
              Rua Doutor Roberto Frias 378, 4200-465 Oporto\\
              Tel.: +351-225081400 \\
              \email{lft@fe.up.pt}
}

\date{Received: date / Accepted: date}

\maketitle

\begin{abstract}
The practicality of a video surveillance system is adversely limited by the amount of queries that can be placed on human resources and their vigilance in response. To transcend this limitation, a major effort under way is to include software that (fully or at least semi) automatically  mines video footage, reducing  the burden imposed to the system.
Herein, we propose a semi-supervised incremental learning framework for evolving visual streams in order to develop a robust and flexible track classification system. 
Our proposed method learns from consecutive batches by updating an ensemble in each time. It tries to strike a balance between performance of the system and amount of data which needs to be labelled. 
As no restriction is considered, the system can address many practical problems in an evolving multi-camera scenario, such as concept drift, class evolution and various length of video streams which have not been addressed before. Experiments were performed on synthetic as well as real-world visual data in  non-stationary environments, showing high accuracy with fairly little human collaboration.                                 
\keywords{Video surveillance \and Parallel streams \and Active learning}
\end{abstract}

\section{Introduction}
\label{intro}
Over the last decades, video surveillance began to spread rapidly, specifically targeted at public areas. Recording for hours, days, and possibly years provides massive amount of information coming from an evolving environment in where traditional learning methods fail to reflect evolution taking place~\citep{Dick03issuesin}. In such environments, the underlying distribution of data changes over time - often referred to  as \emph{concept drift} - either due to intrinsic changes (pose change, movement, etc.), or extrinsic changes (lighting condition, dynamic background, complex object background, changes in camera angle, etc.). Thus, models need to be continually updated to represent the latest concepts. The problem is further aggravated when new objects enter the scene - referred to as \emph{class evolution} in machine learning literature - as new models need to be trained for the novel classes.

\begin{figure}[ht]
\centering
\includegraphics[scale=.5]{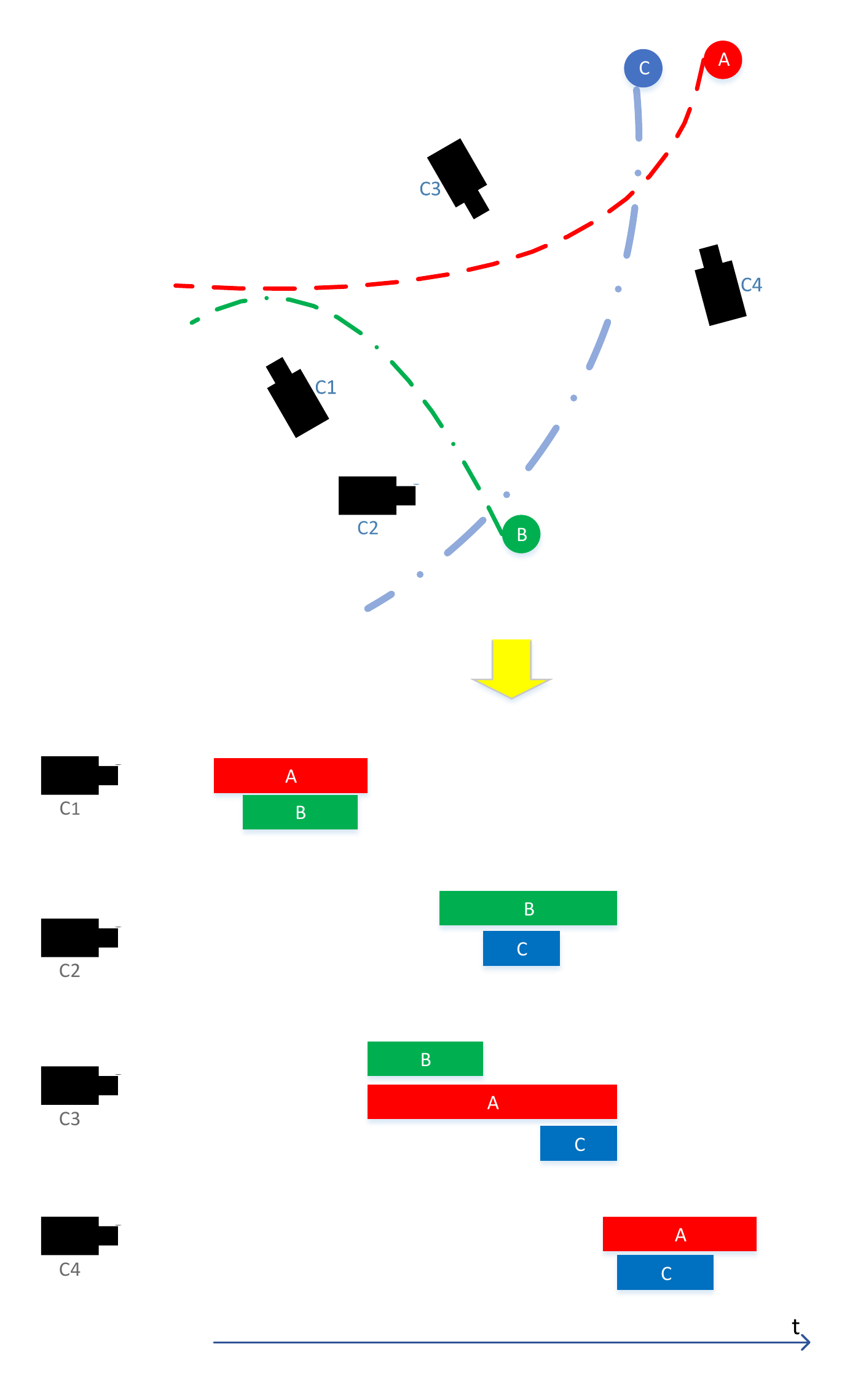} 
\caption{Typical surveillance scenario}
\label{tag01}
\end{figure}

Figure~\ref{tag01} demonstrates a typical surveillance scenario. Depending on the view angle and the quality of the camera, every surveillance camera covers an area called Field of View (FoV). Often the fields of view are disjoint due to budget constraints, whereas they overlap in some scenarios. When entering the scene, the object will enter the coverage area of at least one of the cameras. The surveillance system will have to track that object from the first moment it was captured by a camera and across all cameras whose fields of view overlap the object's path.
In such environments where objects move around and cross the FOV of multiple cameras, it is more than likely to have multiple streams, potentially overlapping in time, recorded at different starting points with various lengths, for the same individual object (Figure~\ref{tag01}). 
However, this type of scenarios is associated with several difficulties. For example, consider the following situation: three different persons are detected by a tracking system in a considered span interval. Person A and person B walk side by side while they are captured by camera 1. Person C enters camera 2 field of view and meets person B. In camera 3, person A and person C start walking side by side. Finally, both are again captured by camera 4, after switching their relative positions. In this simple scenario the typical tracking systems are likely to encounter problems. In fact, mutual occlusion may occur if persons B and C cross. Consequently, their identities can be switched. Moreover, accompanying person A with person B or C, group movement (both are identified as a single object) and prolonged occlusion might occur, which might lead to track loss or mistaken identities~\cite{Teixeira2009}.
Since the cameras are supposed to track all objects in their coverage area, the definition of a global identity for each object is necessary. Multiple appearances of objects captured by the same or by different cameras are identified in the process, allowing also to know the path followed by a given object.
The inference of the identity of the objects in the scene is typically addressed with supervised learning methodologies from labelled training data.
Obtaining labelled instances, which typically needs human annotation, is expensive, time consuming, and impractical for our scenario. To reduce costs of annotation, semi-supervised learning (SSL) approaches have been extensively explored in limited labelled and usually abundant un-labelled data scenarios~\cite{DBLP:conf/pakdd/MasudGKHT10,Cong04semi-supervisedtext,garrette:acl13,Liang05semi-supervisedlearning}; however deploying SSL for evolving visual data in a non-stationary environments (where both concept drift and class evolution are present) is still an unexplored area.
Several researchers have shown that the meticulous selection of instances that need to be labelled (mostly addressed in active learning (AL) strategies)  could lead to better performance with less effort~\cite{Baram:2004:OCA:1005332.1005342, Raghavan06activelearning}.
In this work we address the need for a more general and systematic view of learning in evolving video streams in a multi-camera surveillance scenario. Considerable body of multi-camera surveillance research assume that adjacent camera view have overlap~\cite{Chang01trackingmultiple, Kuo:2010:IAM:1886063.1886093,DBLP:journals/cviu/HamidKHE14,journals/prl/Wang13}, whereas ~\cite{Javed05appearancemodeling,DBLP:conf/cvpr/ShanSK05,DBLP:books/sp/JavedS08, 10.1109/TPAMI.2009.56,Matei:2011:VTA:2191740.2192060} require non-overlapping views. Herein we put forward a framework to learn continuously from parallel video streams with partially labelled data and that allow us to learn novel knowledge, reinforce existing knowledge that is still relevant, and forget what may no longer be relevant. The method made no assumption of overlapping or non-overlapping view. Hence can be applied in either settings. The framework focuses on the classification of multiple video objects being tracked by a video tracking system. The framework receives directly the tracked sequences outputted by the tracking system and maintains a global object identity common to all the cameras in the system. Thus, a suitable outcome of the framework is a timeline graph (such as the one shown in Figure~\ref{tag01}) allocating a stream in each camera for every participant in the system along the time axis for the indicated presence period.

The rest of the paper is organized as follows: next section~\ref{Literature Review} briefly reviews and discusses the limitation of former incremental learning algorithms for visual data. Section~\ref{Never Ending Visual Information Learning} provides an overview of our method. Section~\ref{Experimental Methodology} discusses our experimental methodology. Section~\ref{Results} presents the results of our method as well as some baseline approaches on a variety of synthetic and real datasets. Conclusions and direction for future work are presented in section~\ref{Conclusions}.
\subsection{Literature Review}\label{Literature Review}
\begin{table*}[bp]

       \centering 
       \begin{tabular}{ C{1.6cm}  C{1.6cm} C{1.6cm} C{1.6cm} C{1.6cm} C{1.6cm} C{1.6cm} C{1.6cm} }
    \hline\hline 
    \small Method& \small Parallel Streams& \small Uneven Streams& \small Concept Drift& \small Class  Evolution&  \small Learning & \small Complexity&\small Data \\
    \hline 
\\ \small \cite{Ozawa_incrementallearning, Wu04anonline-optimized} &$\times$&$\times$ &$\surd$& $\times$&\small SL&\small Constrained & MD\\
    \\\small ~\cite{Ackermann2010}&$\times$&$\times$&$\surd$&$\times$&\small SL&\small Unconstrained&MD\\
   \\ \small ~\cite{46}&$\times$&$\times$&$\surd$&$\surd$&\small SL& \small Unconstrained&\small MD\\
   \\ \small ~\cite{DBLP:journals/tkde/MasudGKHT11}&$\times$&$\times$&$\surd$&$\surd$ &\small SSL & \small Constrained&\small MD\\
   \\\small ~\cite{Kolter:2007:DWM:1314498.1390333}&$\times$&$\times$&$\surd$&$\times$&\small Clustering &\small Unconstrained&MD\\
    \\ \small \cite{ BeringerH06, 09, YChen09, ChenZT12}&$\surd$&$\times$&$\surd$&$\surd$&\small Clustering&\small Constrained&1D\\
    \\ \hline
     \hline
    \end{tabular}
 \caption{\label{table:learning assessment}Assessment of learning methods. $\surd$ and $\times$ denote being fit and inappropriate for the purpose, respectively.~``\emph{MD}'' and ``\emph{1D}'' denote multi-dimensional and one-dimensional data. ``SL'' and ``SSL'' indicate Supervised Learning and Semi-Supervised Learning.}
\end{table*} 

Much of the recent history on visual data understanding in general and multi-camera surveillance in particular has focused on building robust models applicable in object detection and tracking scenarios~\cite{DBLP:conf/iccv/LevinVF03, Hebert_2005_4875, Stalder2009BeyondSemi-SupervisedTracking, Teichman-RSS-11, Carvalho2012}. 
Learning changing video streams over time has received much less attention despite the abundance of applications generating this information. Much of the learning literature is concerned with a stationary environment, where fixed and known number of categories to be recognized and enough resources (labelled data, memory and computational power) are available~\cite{Ozawa_incrementallearning, Wu04anonline-optimized}. To get closer to a practical solution, where obtaining labelled instances is an issue, SSL approaches have been deployed. 
various SSL methods have been proposed for video annotation~\cite{Song05semi-automaticvideo,Wang:2009:SKD:1502816.1503021, Xu:2012:SMM:2393347.2396300}. However they have shown promising results for drifting scenarios with pre-determined classes (training data is available for all the classes), but  they cannot address class evolution problem.
In~\cite{Of05personidentification}, the person identification task is posed as a graph-based semi-supervised learning problem, where only a few low quality webcam images are labelled. The framework is able to track various objects in limited drifting environments. 
The classification of objects that have been segmented and tracked without the use of a class-specific tracker, has been addressed with an SSL algorithm in~\cite{Teichman-RSS-11}. Given only three hand-labelled training examples of each class, the algorithm can perform comparably to equivalent fully-supervised methods, but it requires full-length tracks (it is therefore an off-line process) generated by a perfect tracker (each stream represents a single object), which would be challenging for real applications, where multiple streams are available simultaneously. 
Learning from time-changing data has mostly appeared in data mining context and various approaches have been proposed. Ensemble-based approaches constitute a widely popular group of these algorithms to handle concept drift~\cite{Ackermann2010, Kolter:2007:DWM:1314498.1390333} and in some recent works class evolution~\cite{46}, as well. Learn++.NSE~\cite{46} is one of the latest ensemble-based classification methods in literature, that  generates a classifier using each batch of training data and applies a dynamic weighting strategy to define the share of each ensemble in the overall decision. As success is heavily dependent on labelled data, this method would not be applicable in wild scenarios. Masud in~\cite{DBLP:journals/tkde/MasudGKHT11} proposed an online clustering algorithm for single stream that employs an active strategy in order to minimize oracle collaboration.   

Although a considerable body of research has emerged from stream mining, learning from multiple streams in wild environments, that views whole or segments of a stream as a unique element to cluster (or classify), is a less explored area. The methods that have been proposed~\cite{Keogh:2003:NTS:861097.861112, BeringerH06, 09, YChen09,ChenZT12}, require equal length streams coming from a fixed number of sources.  Thus, they would fail to leverage information from time-varying video tracks.
An effective and appropriate algorithm to fit in our scenario is required to: a) learn from multiple streams; b) mine streams with various length and starting points (uneven streams); c) handle the concept drift; d) accommodate new classes; e) deal with partially labelled or unlabelled data; f) be of limited complexity; g) handle multi-dimensional data.

We wrap up our review in Table~\ref{table:learning assessment}, presenting a qualitative look at the extent to which the reviewed methods fulfil the requirements for deploying our scenario. To the best of our knowledge, none of the methods have addressed the problem of learning from multiple streams of visual data. In the next section we discuss our proposed algorithm for stream classification.
\section{Never Ending Visual Information Learning}
\label{Never Ending Visual Information Learning}
In this section we present our Never Ending Visual Information Learning (NEVIL) framework. 
NEVIL is designed for non-stationary data environments in which no labelled data is available but the learning algorithm is able to interactively query the user to obtain the desired outputs at carefully chosen data points. The NEVIL algorithm is an ensemble of classifiers that are incrementally trained (with no access to previous data) on incoming batches of data, and combined with a form of weighted majority voting.
\subsection{Algorithm Description}
\begin{figure*}
\centering
\includegraphics[width=\textwidth]{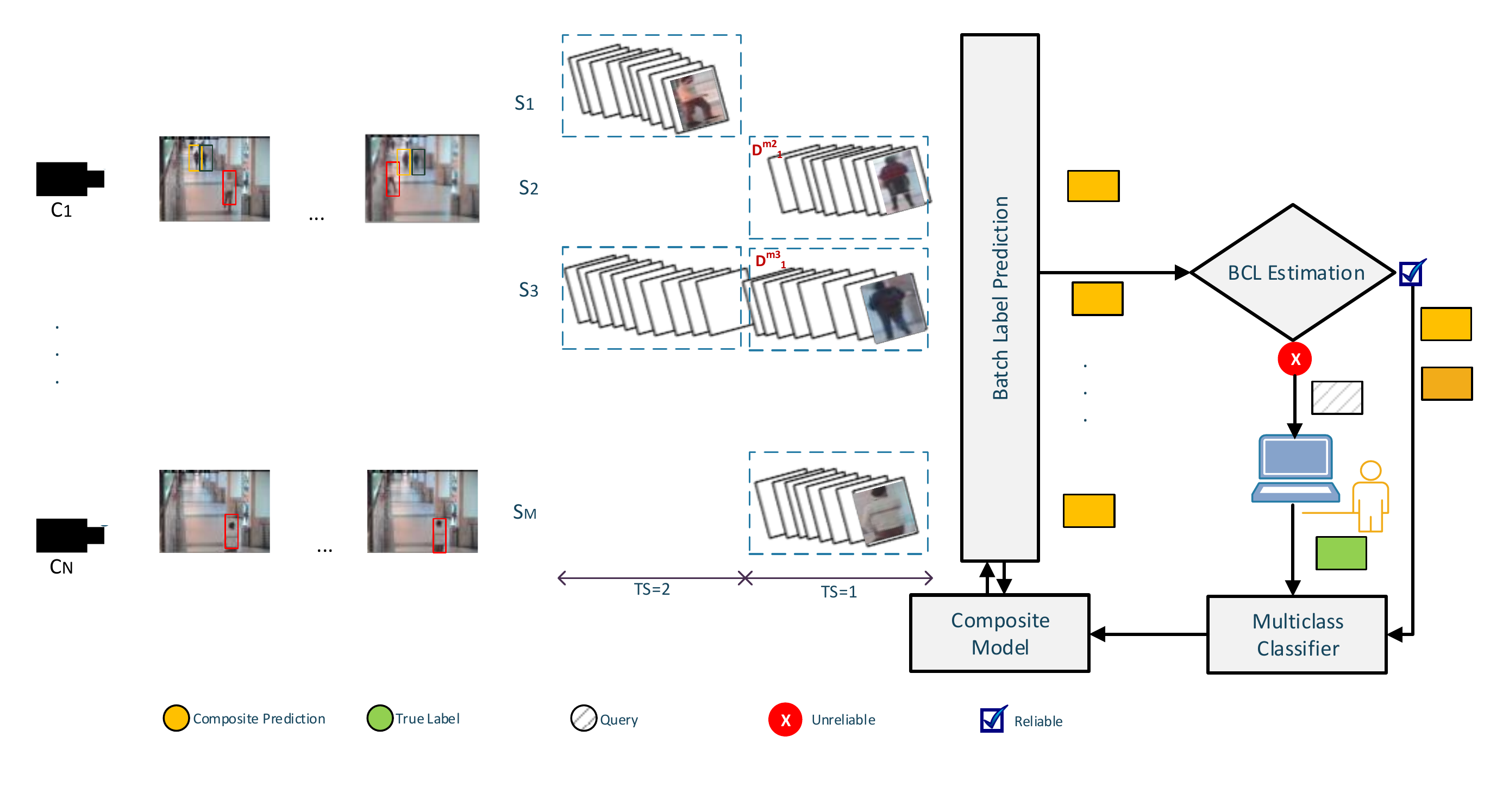}
\caption{NEVIL High-level Overview}
\label{fig:High Level Overview}
\end{figure*}
A high-level sketch of the proposed method is shown in Figure~\ref{fig:High Level Overview}. A typical tracking algorithm analyses sequential video frames and outputs the movement of targets between the frames, generating multiple streams of visual data. Environmental challenges such as varying illumination, lack of contrast, bad positioning of acquisition devices, blurring caused by motion as well as occlusion make data often noisy and/or partially missing. We address these challenges by a batch divisive strategy, as learning from a data batch may reduce the noise and fill the gaps caused by miss-tracking.

The algorithm is provided with a series of data batches ${\cal D}_t^{m_i}$, where $m_i$ is the index of the $i$-th stream present at time slot $t$, $TS_t$, (not all streams are necessarily present). Note that a stream corresponds to a track generated by the tracking system and a single camera can yield multiple streams. A single batch aggregates $B$ frames. The starting time of each stream is potentially different from stream to stream but batches are aligned between streams. Inside each frame the data corresponds to some pre-selected object representation (e.g. bag of words, histogram) and is out of the scope of this paper. 

The ensemble obtained by all models generated up to the current time slot $TS_t$ is named the composite hypothesis $H_{t-1}$. With the arrival of the current data batches ${\cal D}_t^{m_i}, \quad i=1\cdots M$, NEVIL tries to predict the class label for each of the batches in current $TS_t$ based on the probability estimate $p(C_k|{\cal D}_t^{m_i}, H_{t-1})$, where $C_k$ runs over all the class labels observed so far.

This kind of on-line learning approach addressed in this work can suffer if labelling errors accumulate, which is inevitable. Unrelated objects will sooner or later be assigned the same label or different labels will be assigned to different views of the same object. To help mitigate this issue, we allow the system to interact with a human, to help it stay on track. 

Algorithm~\ref{alg:example} outlines our approach. Initially, the composite model is initialized to yield the same probability to every possible class (uniform prior).
When the batches ${\cal D}_1^{m_t}$ in time slot $t$ become available, NEVIL starts with computing the probabilities $p(C_k|{\cal D}_t^{m_i}, H_{t-1})$ for each batch  ${\cal D}_t^{m_i}$ in the time slot.
Once $p(C_k|{\cal D}_t^{m_i}, H_{t-1})$ is obtained, a batch confidence label (BCL) is estimated; if BCL is high enough (above a prespecified threshold), the predicted label $$\arg \max_{C_k} p(C_k|{\cal D}_t^{m_i}, H_{t-1})$$ is accepted as correct, otherwise the user is requested to label the data batch. The labelled batches (either automatically or manually) are used to generate a new multiclass classifier that is integrated in the composite model, yielding $H_{t}$.

Four tasks need now to be detailed: a) the batch label prediction (by the composite model); b) the BCL estimation; c) the multiclass classifier design in current time slot; d) the composite model structure and update. 
\begin{algorithm}[tb]
   \caption{NEVIL}
   \label{alg:example}
\begin{algorithmic}
  \STATE {Input:}  ${\cal D}_{t}^{m_i},i={1,...,M}$
  \STATE {$W_0$ $\leftarrow$ $\frac{1}{k}$}
  \STATE { $H_0$ $\leftarrow$ $W_0$}
  \WHILE{${\cal D}_{t}$ is $True$}
  \STATE {\bfseries Batch label prediction~(Section~\ref{Batch Label Prediction})} 
  \STATE {$p(C_k|{\cal D}_{t}^{m_i})$ $\leftarrow$ $({\cal D}_{t}^{m_i}, H_{t-1})$}
   \STATE {\bfseries Batch Confidence Level Estimation~(Section~\ref{The Batch Confidence Level Estimation})}  
   \STATE {$BCL$ $\leftarrow$ $p(C_k|{\cal D}_{t}^{m_i}, H_{t-1})$}
     \STATE {\bfseries Multiclass classifier design~(Section~\ref{Multiclass Classifier})}  
  \STATE {$h_t$ $\leftarrow$ ${\cal D}_{t}$}
     \STATE {\bfseries Composite model structure and update~(Section~\ref{The Composite Model Structure and Update})}  
  \STATE {$H_t$ $\leftarrow$ $(h_t,~H_{t-1},~W_{t})$}
  \ENDWHILE
\end{algorithmic}
\end{algorithm}

\subsubsection{Batch Label Prediction}
\label{Batch Label Prediction}
A batch ${\cal D}_t^{m_t}$ is a temporal sequence of frames ${\cal D}_{t,f}^{m_t}$, where $f$ runs over $1$ to the batch size $B$.
The composite model, $H_{t-1}$, can be used to predict directly $p(C_k|{\cal D}_{t,f}^{m_i}, H_{t-1})$ but not $p(C_k|{\cal D}_{t}^{m_i}, H_{t-1}).$
The batch (multiframe) Bayesian inference requires conditional independence 
$$
\begin{array}{l}
p({\cal D}_{t}^{m_i}| C_k, H_{t-1}) = \\
p({\cal D}_{t,1}^{m_i},\cdots, {\cal D}_{t,B}^{m_i}| C_k, H_{t-1}) = \\
p({\cal D}_{t,1}^{m_i}| C_k, H_{t-1})\cdots p({\cal D}_{t,B}^{m_i}| C_k, H_{t-1})=\\
\prod_{j=1}^B p({\cal D}_{t,j}^{m_i}| C_k, H_{t-1})
\end{array}
$$
From there, and assuming equal prior probabilities, it is trivial to conclude that 
\begin{equation}
\label{eq:geometric}
p(C_k|{\cal D}_{t}^{m_i}, H_{t-1}) = Z \prod_{j=1}^B p(C_k|{\cal D}_{t,j}^{m_i}, H_{t-1}),
\end{equation}
where $Z$ is a normalization constant.
In practice, products of many small probabilities can lead to numerical underflow problems, and so it is convenient to work with the logarithm of the distribution.
The logarithm is a monotonic function, so that if $p(C_k|{\cal D}_{t}^{m_i}, H_{t-1}) > p(C_\ell|{\cal D}_{t}^{m_i}, H_{t-1})$ then $$\log p(C_k|{\cal D}_{t}^{m_i}, H_{t-1}) > \log p(C_\ell|{\cal D}_{t}^{m_i}, H_{t-1}).$$ 
Then we can rewrite the decision as choosing the class that maximizes
\begin{equation}
\label{eq:loggeometric}
\log p(C_k|{\cal D}_{t}^{m_i}, H_{t-1}) = \log Z + \sum_{j=1}^B \log p(C_k|{\cal D}_{t,j}^{m_i}, H_{t-1})
\end{equation}

The batch label prediction can also be analysed as a problem of combining information from multiple ($B$) classification decisions. Considering that, per frame, the composite model produces approximations to the a posteriori probabilities of each class, different combination rules can be considered to build the batch prediction from the individual frame predictions~\cite{Alexandre20011283,Kittler1998}. While Equation~\eqref{eq:geometric} turns out to be the product rule (or geometric mean), the sum rule (or arithmetic mean) is also often preferred:
\begin{equation}
\label{eq:arithmetic}
p(C_k|{\cal D}_{t}^{m_i}, H_{t-1}) = Z\sum_{j=1}^B p(C_k|{\cal D}_{t,j}^{m_i}, H_{t-1})
\end{equation}

In fact some authors have shown that the arithmetic mean outperforms the geometric mean in the presence of strong noise~\cite{Alexandre20011283,Kittler1998}.
Experimentally, we will compare both options.

\subsubsection{The Batch Confidence Level Estimation}
\label{The Batch Confidence Level Estimation}
Having predicted a class label for a data batch, one needs to decide if the automatic prediction is reliable and accepted or rather a manual labelling be requested.

Various criteria have been introduced as uncertainty measures in literature for a probabilistic framework~\cite{Settles10activelearning}. 
Perhaps the simplest and most commonly used criterion relies on the probability of the most confident class, defining the confidence level as
\begin{equation}
\label{eq:BCL1}
\max_{C_k} p(C_k|{\cal D}_{t}^{m_i}, H_{t-1}).
\end{equation}
However, this criterion only considers information about the most probable label. Thus, it effectively ``throws away'' information about the remaining label distribution~\cite{Settles10activelearning}.

To correct for this, an option is to adopt a margin confidence measure based on the first and second most probable class labels under the model:
\begin{equation}
\label{eq:BCL2}
p(C^*|{\cal D}_{t}^{m_i}, H_{t-1})-p(C_*|{\cal D}_{t}^{m_i}, H_{t-1}),
\end{equation}
where $C^*$ and $C_*$ are the first and second most probable class labels, respectively.
Intuitively, batches with large margins are easy, since the classifier has little doubt in differentiating between the two most likely class labels. Batches with small margins are more ambiguous, thus knowing the true label would help the model discriminate more effectively between them~\cite{Settles10activelearning}.

Note that while the estimation of the wining class for batch label prediction requires only the comparison of the relative values as given by~\eqref{eq:geometric}, \eqref{eq:loggeometric} or~\eqref{eq:arithmetic}, both approaches \eqref{eq:BCL1} and \eqref{eq:BCL2} for the confidence level require the exact computation of the a posteriori probabilities of the classes. This involves computing the normalizing constant associated with~\eqref{eq:geometric} or \eqref{eq:arithmetic}, which is specially unstable for~\eqref{eq:geometric}.

We therefore put forward variants of the two previous measures.
As an alternative to the margin confidence measure~\eqref{eq:BCL2}, we base the confidence level on the {\em ratio} of the first and second most probable class labels: 
\begin{equation}
\label{eq:BCL3}
BCL=p(C^*|{\cal D}_{t}^{m_i}, H_{t-1})/p(C_*|{\cal D}_{t}^{m_i}, H_{t-1}),
\end{equation}
which can be directly applied for the sum rule or modified to $\log p(C^*|{\cal D}_{t}^{m_i}, H_{t-1}) - \log p(C_*|{\cal D}_{t}^{m_i}, H_{t-1})$ for the product rule.
Either way, we eliminate the issue with the normalization constant.

To come up with an alternative to the most confident class measure, we write the decision as
 \begin{equation}
\label{eq:BCL4}
\max p_k = \frac{\prod_{j=1}^B p_{k,j}}{\sum_k \prod_{j=1}^B p_{k,j}} \gtrless T,
\end{equation}
where we introduced the following simplifications in notation: $p_k = p(C_k|{\cal D}_{t}^{m_i}, H_{t-1})$ and $p_{k,j}=p(C_k|{\cal D}_{t,j}^{m_i}, H_{t-1})$. 
The comparison in Eq.~\eqref{eq:BCL4} can be rewritten as 
 \begin{equation}
\label{eq:BCL5}
(1-T)\prod_{j=1}^B p_{k^*,j} \gtrless T \sum_{k, k\not = k^*} \prod_{j=1}^B p_{k,j},
\end{equation}
where $k^*=\arg\max_k p_k$. 
Since we cannot work directly with the $\log$ of~\eqref{eq:BCL5} due to the sum in the denominator, we introduce the simplification of binarizing the classification in each frame, defining $\bar{p}_{k^*,j}=\sum_{k, k\not = k^*}p_{k,j}=1-p_{k^*,j}$. 

Accepting the strong assumption of independence for the aggregated class, then  $$\bar{p}_{k^*}= \prod_{j=1}^B \bar{p}_{k^*,j}.$$ This ends up in exchanging the order of the sum and product in the right hand side of~\eqref{eq:BCL5}, which can now be rewritten as 
\begin{equation}
\label{eq:BCL6}
(1-T)\prod_{j=1}^B p_{k^*,j} \gtrless T \prod_{j=1}^B \bar{p}_{k^*,j}.
\end{equation}
Now it is a trivial process to apply the $\log$ to obtain a stable decision:
\begin{equation}
\label{eq:BCL7}
\sum_{j=1}^B \log p_{k^*,j} \gtrless  S + \sum_{j=1}^B \log \bar{p}_{k^*,j},
\end{equation}
where $S=\log T - \log (1-T)$.

\begin{figure*}
        \centering
        \begin{subfigure}[b]{0.2\textwidth}
                \centering
                \includegraphics[width=\textwidth]{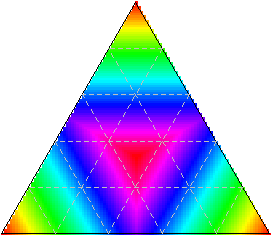}
                \caption{Most confident class measure.}
                \label{fig:LC}
        \end{subfigure}%
        ~  
        \begin{subfigure}[b]{0.2\textwidth}
                \centering
                \includegraphics[width=\textwidth]{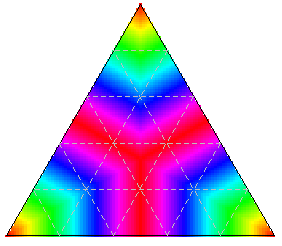}
                \caption{Margin confidence measure.}
                \label{fig:Margin}
        \end{subfigure}
         \begin{subfigure}[b]{0.2\textwidth}
                \centering
                \includegraphics[width=\textwidth]{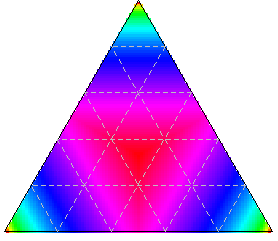}
                \caption{Modified most confident class measure.}
                \label{fig:AdaptedLC}
        \end{subfigure}%
        ~  
        \begin{subfigure}[b]{0.2\textwidth}
                \centering
                \includegraphics[width=\textwidth]{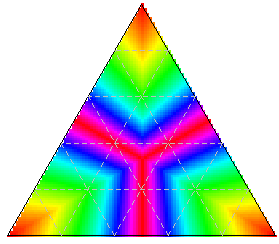}
                \caption{Modified margin confidence measure.}
                \label{fig:AdaptedMargin}
        \end{subfigure}
        ~
        \begin{subfigure}[b]{0.12\textwidth}
                \centering
                \includegraphics[scale=0.5]{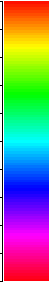}
                \label{fig:colorbar}
        \end{subfigure}
        \caption{Heatmaps illustrating the behavior of the reliability measures in a three-label classification problem.}\label{fig:BCL}
\end{figure*}

Figure~\ref{fig:BCL} highlights the characteristics of the four confidence measures by a ternary plot (where every corner indicates a class). This plot graphically depicts the ratios of the three variables (herein, occurrence of each class) as positions in an equilateral triangle. The probability of each class is $1$ in its corner of the triangle. Moving inside triangle, the percentage of a specific class decreases linearly with increasing distance from the corner till dropping to 0 at the line opposite it.
A rainbow-like color pattern shows the informativeness of different composition of three classes. For all methods, the least reliable batch would lie at the center of triangle, where the posterior label distribution is uniform and thus the least certain under the ensemble. Similarly, the most informative batch lies at the corners where one of the classes has the highest possible probability. 
 
\subsubsection{Multiclass Classifier}
\label{Multiclass Classifier}
At time slot $t$, we obtain a new set of batches that are automatically or manually labelled. We assume all the frames belonging to a batch are from the same object (and the underlying tracking system does not mix identities in the time slot period) and therefore the frames inside a batch correspond to observations of the same class. Consider that to the $M$ batches in current time slot correspond $L<M$ labels (some batches can have the same label). 
We need the design a classifier that can approximate the a posteriori probability function $p(c_k|{\cal D}_{t,f}^{m_i})$, which gives the probability of the frame belonging to a given class $c_k$, given that ${\cal D}_{t,f}^{m_i}$ was observed.

A standard way to address this problem is to apply \emph{discriminative approaches} which predict the conditional probability directly. As an alternative, \emph{generative approaches} find the joint distribution $p({\cal D}_{t,f}^{m_i}, c_k)$ and then use Bayes' rule to form the conditional distribution from the generative model.
A third option is to find a function $f({\cal D}_{t,f}^{m_i})$, called a discriminant function, which maps each input ${\cal D}_{t,f}^{m_i}$ directly onto a class label. In this case, and although probabilities play no role in the design of the discriminant function, it is still possible to get estimated for the conditional probabilities~\cite{Bishop:2006:PRM:1162264}. Each approach has its relative merits and we evaluate experimentally instantiations of each.

One of the challenges we need to handle in a practical scenario is when in a time slot all the batches have the same label (automatically or manually assigned). In these TSs the training of a multiclass classifier is not possible. We resort to one-class classifiers for these time slots, also known as unary classification, to distinguish the single class present in the training set (the batches in the time slot) from all other possible classes~\cite{Khan2010}.
\subsubsection{The Composite Model Structure and Update}
\label{The Composite Model Structure and Update}
The composite model $H_t$ in the NEVIL framework is an ensemble of classifiers $h_t$ that are incrementally trained (with no access to previous data) on incoming time slots of data as described previously. The individual models $h_t$ are combined using a weighted majority voting, where the weights are dynamically updated with respect to the classifiers' time of design.

The prediction outputted by the composite model $H_t$ for a given frame ${\cal D}_{t,f}^{m_i}$ is $$p(C_k|{\cal D}_{t,f}^{m_i}, H_{t}) = \sum_{\ell=1}^t W_\ell^t h_\ell(C_K|{\cal D}_{t,f}^{m_i}),$$
where $h_\ell(.)$ is the multiclass classifier trained at TS $\ell$, $W_\ell^t$ is the weight assigned to classifier $\ell$, adjusted for time $t$.

The weights are updated and normalised at each time slot and chosen to give more credit to more recent knowledge. The weights are chosen from a geometric series ${\frac{1}{p^t},...,\frac{1}{p^2},\frac{1}{p}}$, normalised by the sum of the series to provide proper probability estimates:
$$W_\ell^t = \frac{\frac{1}{p^{(t-\ell+1)}}}{\sum_{j=1}^t \frac{1}{p^j}}$$

\section{Experimental Methodology}
\label{Experimental Methodology}
A series of experiments were conducted to explore the capabilities of the proposed framework. In order to study the behaviour of the system facing various conditions, we generated multiple synthetic streams that were organized in different scenarios. We also tested the NEVIL framework with real video data (see section~\ref{Datasets}). 

\begin{table*}[bp]
\caption{Parametric Equations for classes of MS dataset }
\label{table:parameters}
\scalebox{0.77}{
\begin{tabular}{lllllllllllllllll}
\hline
 \toprule 
Drift Rate & \multicolumn{4}{c}{\bf {C1}}  & \multicolumn{4}{c}{\bf C2} & \multicolumn{4}{c}{\bf C3} & \multicolumn{4}{c}{\bf C4} \\

\small  & $\mu_{x} $&$ \mu_{y}$ &  $\delta_{x}$ & $\delta_{y}$
    & $ \mu_{x} $&$ \mu_{y}$ &  $\delta_{x}$ & $\delta_{y}$
     & $ \mu_{x} $& $\mu_{y}$ & $ \delta_{x}$ & $\delta_{y}$
      &$  \mu_{x} $& $\mu_{y}$ & $ \delta_{x}$ &$ \delta_{y} $ \\
\midrule

\small $0<r<0.25$& 2&5 &$0.5r$&$0.5+2r$
&$5-5r$& 8& $3-10r$&1
 &$5-5r$&2&$0.5+10r$ &0.5
 &8& $8+15r$&0.5&0.5\\
 \hline
 \small $0.25<r<0.5$& $-$&$-$ &$-$&$-$
&15&$-1+5r$&1&$2+3r$
 &$1-4r$&2&0.5 &$3-4r$
 & $5-5r$&13& $0.25+4r$& $0.5+4r$\\
 \hline
 \small $0.5<r<0.75$& 10&$-10+5r$ &$1$&$2-r$
&17&$2+5r$&0.25&0.15
 &-1&$-2-4r$&0.25 &0.15
 & $-$&$-$ &$-$&$-$\\
 \hline
 \small $0.75<r<1$& $-$&$-$ &$-$&$-$
&20&$4r$&1&2
 &-7&$-5-r$& $7+4r$& $1+4r$
 & $-$&$-$ &$-$&$-$\\
 
 \hline
 \toprule 
Drift Rate & \multicolumn{4}{c}{\bf C5} & \multicolumn{4}{c}{\bf C6} & \multicolumn{4}{c}{\bf C7} \\

\small  & $\mu_{x} $&$ \mu_{y}$ &  $\delta_{x}$ & $\delta_{y}$
    & $ \mu_{x} $&$ \mu_{y}$ &  $\delta_{x}$ & $\delta_{y}$
     & $ \mu_{x} $& $\mu_{y}$ & $ \delta_{x}$ & $\delta_{y}$ \\
\midrule

\small $0<r<0.25$& 12&15 &2&$2+2r$
&-15& $-5+15r$& 1&$2+3r$
 &10&$5r$&0.5&$2+3r$ \\
 \hline
 \small $0.75<r<1$&  & $-$&$-$ &$-$&$-$
  & $-$&$-$ &$-$&$-$
  $-10$&$-1+5r$ &$r$&$2+3r$\\
\bottomrule 
\end{tabular}
}
\end{table*}
\subsection{Datasets}
\label{Datasets}
In order to explore the properties of the proposed framework, we evaluated it on multiple datasets covering various possible scenarios in a multi-camera surveillance system.

We conducted our experiments on synthetic as well as real datasets.
The synthetic dataset is generated in the form of $(X,y)$, where $X$ is a 2-dimensional feature vector, drawn from a Gaussian distribution~N($\mu_{X}$, $\delta_{X}$), and $y$ is the class label. 

Since in real applications visual data may suffer from both gradual and abrupt drift, we tried to simulate both situations in our streams by changing $\mu_{X}$ and $\delta_{X}$ in the parametric equations; Table~\ref{table:parameters} presents these parametric equations. In this experiment, we generated 7 classes (${C1,C2,...,C7}$); for some $(C5,C6)$ data changes gradually while others also experience one $(C1, C4, C7)$, or three $(C2, C3)$ dramatic drifts. This process is similar to the one used in~\cite{46}.

\begin{figure*} 
\centering
$\begin{tabular}{c c}
\fbox{\includegraphics[scale=0.552]{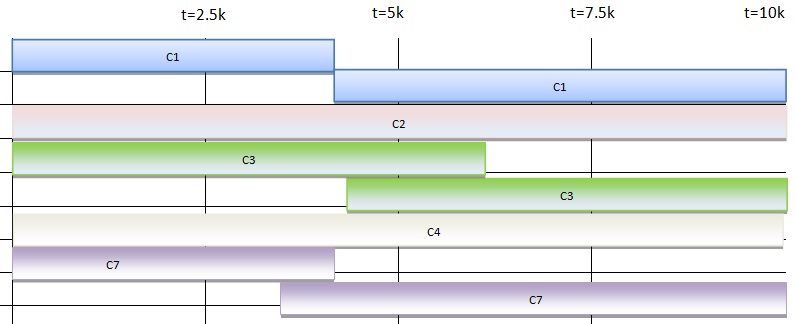}}& \emph{Scenario} \MakeUppercase{\romannumeral 1}\\
\fbox{\includegraphics[scale=0.555]{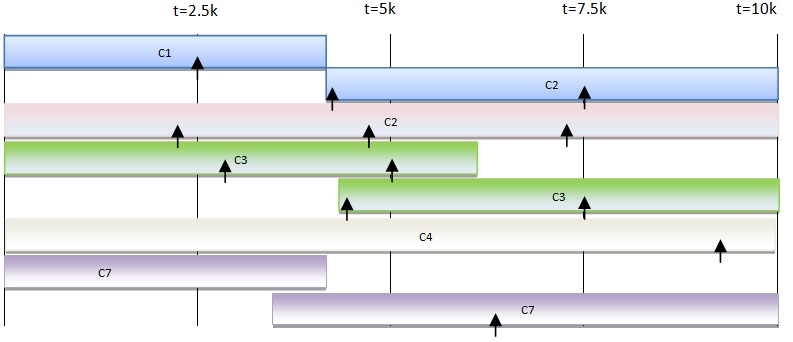}}& \emph{Scenario} \MakeUppercase{\romannumeral 2}\\
\fbox{\includegraphics[scale=0.564]{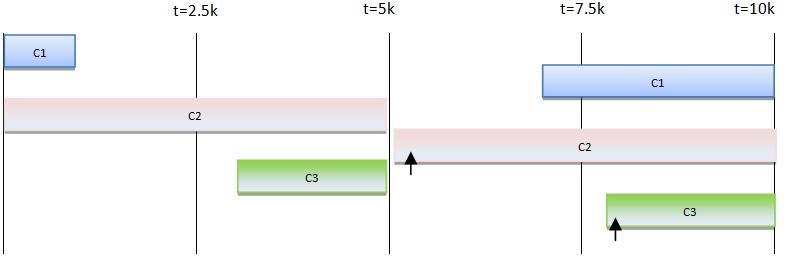}}& \emph{Scenario} \MakeUppercase{\romannumeral 3} \\
\fbox{\includegraphics[scale=0.594]{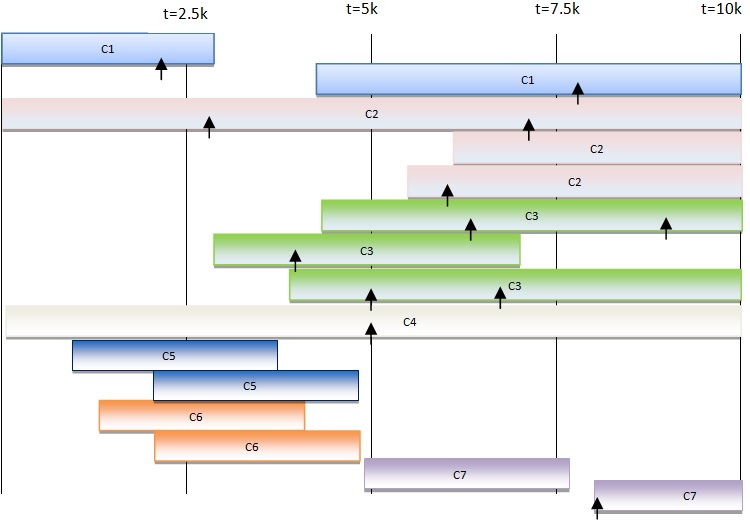}}& \emph{Scenario} \MakeUppercase{\romannumeral 4}\\
\end{tabular}$
\caption{\label{fig:SynScen} \sf Scenarios in MS dataset. ~The sign~\protect\includegraphics[scale=0.5]{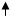} denotes the occurrence of an abrupt drift in the nature of data.}
 \end{figure*}

\begin{figure}[h]
\begin{center}
$\begin{array}{ccccc}
\includegraphics[height=0.7in]{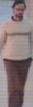} &
\includegraphics[height=0.7in]{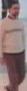}&
\includegraphics[height=0.7in]{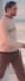} &
\includegraphics[height=0.7in]{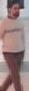}&
\includegraphics[height=0.7in]{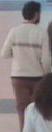} 
\end{array}$
\caption{\label{fig:Div}An example of diversity in appearance}
\end{center}
\end{figure}

\begin{figure*} 
\centering
\setlength\fboxrule{0.5pt}
\fbox{\includegraphics[scale=0.58]{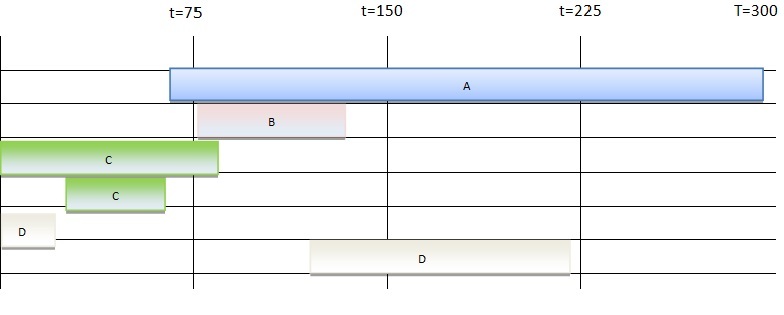}} 
\caption{\label{fig:caviars}The EnterExitCrossingPaths1 scenario in the CAVIAR dataset.}
\end{figure*}

The dataset was organized in 4 different scenarios with different levels of complexity, including streams with gradual drift, sudden drift, re-appearance of objects and non-stationary environments where we have abrupt class and concept drift. Each scenario includes:
\begin{itemize}
\item \emph{Scenario} \MakeUppercase{\romannumeral 1}: gradually drifting streams of 5 classes. 
\item \emph{Scenario} \MakeUppercase{\romannumeral 2}: streams with abrupt drifts of 5 classes. 
\item  \emph{Scenario} \MakeUppercase{\romannumeral 3} : re-appearance of objects.
\item   \emph{Scenario} \MakeUppercase{\romannumeral 4}: a non-stationary environment with class evolution as well as concept drift.
\end{itemize}
These scenarios are depicted in Fig.~\ref{fig:SynScen}.
Besides synthetic datasets, we run our experiments on a number of CAVIAR video clips~\cite{CAVIARdataset} including: OneLeave ShopReenter1, Enter ExitCrossingPaths1, OneShopOneWait1, OneStop Enter2 and WalkBy Shop1front. Due to the presence of different perspectives of the same person, streams are drifting in time (see Fig.~\ref{fig:Div}). These sequences present challenging situations with cluttered scenes, high rates of occlusion, different illumination conditions as well as different scales of the person being captured. We employ an automatic tracking approach~\cite{DBLP:conf/icip/TeixeiraCCC12} to track objects in the scene and generate streams of bounding boxes, which define the tracked objects' positions. As the method may fail to perfectly track the targets, a stream often includes frames of distinct objects.
An hierarchical bag-of-visterms method is applied to represent the tracked objects, resulting in a descriptor vector of size 11110 for each frame (refer to~\cite{Teixeira2009}~for additional details). In order to avoid the curse of dimensionality that system may suffer from, Principle Component Analysis (PCA) is applied to the full set of descriptor features as a pre-processing step. Hence, the number of features in each stream is reduced to 85 dimensions. As an explanatory sample, figure~\ref{fig:caviars} depicts the streams in the EnterExitCrossingPaths1 scenario. 

\subsection{Instantiation of Classifiers}
\label{Instantiation of Classifiers}
In Section~\ref{Multiclass Classifier}, we identified three approaches that have been applied in the literature to obtain the posterior probability. A set of experiments were conducted in order to study the behaviour of our framework employing instances of each option. We chose the following methods:
\emph{Gaussian Mixture Models (GMM)} and \emph{Naive Bayes} as \emph{generative approaches}, \emph{Support Vector Machines (SVM)}~\cite{CC01a} as one of the most popular \emph{discriminant function} and \emph{logistic regression}~\cite{DBLP:journals/jmlr/FanCHWL08} as a member of \emph{discriminative approaches} family. 

Designing a classifier for time slots where batches constitute different labels is quite straightforward. The challenging situation arises when we need to do unary classification.  As we employed various approaches with specific characteristics, different strategies are proposed to handle this situation.

Single-class SVM classifies each frame as completely similar or different from given class, whereas generative approaches (GMM and Naive Bayes) provide the probabilistic estimation. 

To the extent of our knowledge, using logistic regression in unary problems is an unexplored topic; existing methods need data generated by at least two classes in order to make the prediction. Therefore, we keep the batches from the last multi-class time slot and combine them with the uni-class time slot to build the training set.
\subsection{Evaluation Criteria}
\label{Evaluation Criteria}
Active learning aims to achieve high accuracy using as little annotation effort as possible. Thus, a trade-off between accuracy and proportion of labelled data can be considered as one of the most informative measures.\\
Let $N$ denote the total number of batches, $MC$ refer to misclassified batches, then the accuracy of the system in a given time slot is formulated as:
 \begin{eqnarray}
\label{eq:Accuracy}
 Accuracy=\frac{N-\# MC}{N}
  \end{eqnarray}
The total accuracy of a system over a period of time is derived from the mean accuracy of all the time slots.
 
Assume $MLB$ and $TB$ denote the manually labelled batches and all the batches available during a period (includes one or more time slots), respectively.\\ The \emph{Annotation Effort} is formulated as:
\begin{eqnarray}
\label{eq:Annotation}
 \textrm{\it{Annotation effort}}=\frac{\# MLB}{\# TB }
  \end{eqnarray}
  One expects that the accuracy increases with the increase of annotation effort.
    \subsection{Baseline Methods}\label{Baseline Method}
To the best of our knowledge, there is no method that mines multi-dimensional parallel streams in such a non-stationary environment, where the number and length of streams vary greatly. Therefore, we compare our framework with three baseline approaches: 
\begin{itemize}
\item Passive Learning: The first half of all the batches are submitted to the oracle for labelling. Once the labelled set is obtained, a classifier is trained and applied to classify the other half of stream. This method is far from a real online active learning strategy, as it needs complete data available.
For datasets in which there is no dramatic distribution evolution between first and second half, we expect that it provides an upper bound to be compared with our method. 
\item Even/Odd Learning: As an on-line baseline, for a given stream, batches are marked alternately with odd and even integers, where odd batches are kept in a buffer with their true labels. At each time slot, a model is re-trained using the buffer. We then use this model to classify even batches. Therefore, we may partly follow the distribution changes in this setting leading to better performance than Passive Learning. However, we need to keep all the odd batches, which is far from a practical solution in an on-line scenario.
\item  Unwise active learning: We use an unwise version of the original framework as a baseline, where all the batches occurred before initiation time ($t_{int}$) would be annotated.
For $t>t_{int}$, NEVIL computes the probabilities of known classes.
Once $p(C_k|{\cal D}_t^{m_i}, h_{t_{int}})$ are obtained, a batch confidence label (BCL) is estimated; if BCL is high (above a pre-defined threshold), the predicted label $$\arg \max_{C_k} p(C_k|{\cal D}_t^{m_i}, h_{t_{int}})$$ is accepted as correct label of the batch, otherwise the user is requested to label the batch. The method is summarized in Algorithm~\ref{alg:Unwise active learning}. Despite meticulous selection of queries, as the model is not updated, the algorithm may establish a lower 
bound the level of performance that can be expected 
in an evaluation.
\end{itemize}
\begin{algorithm}[tb]
   \caption{Unwise active learning}
   \label{alg:Unwise active learning}
\begin{algorithmic}
  \STATE {Input:}  ${\cal D}_{t}^{m_i},i={1,...,M}$
  \STATE $h \leftarrow empty$
  \WHILE{${\cal D}_{t}$ is $True$}
\IF{$t>t_{int}$} 
  \STATE {\bfseries Batch label prediction} 
  \STATE {$p(C_k|{\cal D}_{t}^{m_i})$ $\leftarrow$ $({\cal D}_{t}^{m_i}, h_{int})$}
   \STATE {\bfseries Batch Confidence Level Estimation}  
   \STATE {$BCL$ $\leftarrow$ $p(C_k|{\cal D}_{t}^{m_i}, h_{int})$}
   \ELSE
     \STATE {\bfseries Multiclass classifier design}  
  \STATE {$h_{int}$ $\leftarrow$ ${\cal D}_{t}$}
       \ENDIF
  \ENDWHILE
\end{algorithmic}
\end{algorithm}    

\section{Results}\label{Results}
Firstly, multiple tests were run to determine the optimal batch size for each dataset to be explored. Batch size was varied between 1\% to 50\% of the size of the longest stream available in each dataset. Experiments were repeated for 50 equally spaced values in that range.
The optimal batch size varies and is influenced by the characteristics of the streams present in each dataset. Optimal batch sizes have been observed to range between 30 and 35 for real video streams and between 200 and 300 for synthetic sequences.


Table~\ref{table:baseline results} provides a summary of the performance of \emph{Passive Learning} and \emph{Even/Odd Learning} using various classifiers on all datasets mentioned in Section~\ref{Datasets}. 
Since different classifiers provide varying performances on different datasets, the need for a procedure that carefully assesses algorithms seems inevitable. We applied Friedman test~\cite{Demsar:2006:SCC:1248547.1248548} that provides a non-parametric rank based statistical significance test. This test is similar to parametric repeated measures ANOVA, which tests if there is a significant difference between the rank of different treatments across multiple attempts. When the test runs over all the datasets shows that null hypothesis is verified which means that type of the classifier has no significant effect on the overall performance of baseline method in real applications. However, the test shows that Logistic Regression has yielded weak results for synthetic data in both learning methods. When we perceive the superior learners based on the mean rank for various scenarios, generative approaches perform fairly better in the synthetic datasets, while discriminative methods win for real video clips. Since the dimension of real data is large, while synthetic data is generated in 2D space, these results also emphasizes the difficulties that generative models face in high-dimensional spaces.
As mentioned in~\ref{Baseline Method}, we expect better or equal results from \emph{Even/Odd Learning} than \emph{Passive Learning} which is the case in all the settings applied discriminative approaches as well as almost all used generative methods. Unexpected behaviour of generative methods when applies on \emph{OneShopOneWait1} dataset can be explained by high bias of these methods when trying to model such complex data.
\begin{table*}[bp]
       \centering 
       \caption{Comparison of baseline approaches on multiple datasets} 
\begin{tabular}{lccc }
& & \multicolumn{2}{ c }{Accuracy (\%)}\\
\hline
\hline
Dataset& Multiple Classifier& Passive Learning&Even/odd Learning\\ \hline
\multirow{4}{*}{Scenario\MakeUppercase{\romannumeral 1}} & SVM & 97.39&97.19 \\
 & GMM &79.61&79.45 \\
 & Naive bayes & 79.61&79.45 \\
 & Logistic Regression &18.44& 18.21 \\ \hline
\multirow{3}{*}{Scenario\MakeUppercase{\romannumeral 2}} & SVM &66.32&72.25 \\
 & GMM &79.60&79.45 \\
 & Naive bayes &79.60&79.45 \\
 & Logistic Regression & 40.85&35 \\ \hline
 \multirow{3}{*}{Scenario\MakeUppercase{\romannumeral 3}} & SVM &74.70&76.65  \\
 & GMM &78.37&78.02 \\
 & Naive bayes & 78.37&78.02\\
 & Logistic Regression &62.18 &62.64 \\ \hline
 \multirow{3}{*}{Scenario\MakeUppercase{\romannumeral 4}} & SVM &80.05& 78.61  \\
 & GMM &81.81&81.87 \\
 & Naive bayes &81.81&81.87\\
 & Logistic Regression & 45.51& 40.65 \\ \hline
 \toprule
 \multirow{3}{*}{EnterExitCrossingPaths1} & SVM & 89.28&93.7  \\
 & GMM &66.45& 75.16\\
 & Naive bayes& 66.45&75.16 \\
 & Logistic Regression & 80.12&79.86\\ 
 \hline
 \multirow{3}{*}{OneLeaveShopReenter1} & SVM &63.74& 100 \\
 & GMM &64.06&61.49 \\
 & Naive bayes & 64.06&61.49 \\
 & Logistic Regression &92.18 &97.86 \\ 
 \hline
 \multirow{3}{*}{OneShopOneWait1} & SVM &80.24 & 95.79 \\
 & GMM &92.33&52.88 \\
 & Naive bayes &92.33 &52.88 \\
  & Logistic Regression & 81.35& 93.42\\ \hline
    \multirow{3}{*}{OneStopEnter2} & SVM & 83.56&98.63  \\
 & GMM &76.73&75.28 \\
 & Naive bayes & 76.73&75.28 \\
 & Logistic Regression &81.36&93.02 \\ \hline
\multirow{3}{*}{WalkByShop1front} & SVM &92.32 &97.58 \\
 & GMM &91.50&90.93 \\
 & Naive bayes& 91.50&90.93 \\
 & Logistic Regression &89.04 &96.07 \\ \hline
 \multirow{3}{*}{OneStopMoveEnter1} & SVM &62.47&79.40 \\
 & GMM &90.99 &90.93 \\
 & Naive bayes &90.99 &90.93 \\
 & Logistic Regression &56.25 &73.76\\ \hline
\hline
 \label{table:baseline results}
\end{tabular}
\end{table*}

\begin{figure}
        \centering
        \begin{subfigure}[b]{0.25\textwidth}
                \centering
                \includegraphics[width=\textwidth]{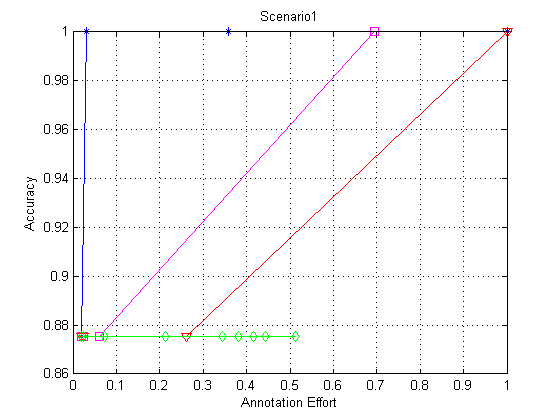}
                \caption{\tiny Naive Classifier}
                \label{fig:naive-scenario1}
        \end{subfigure}%
        ~  
        \begin{subfigure}[b]{0.25\textwidth}
                \centering
                \includegraphics[width=\textwidth]{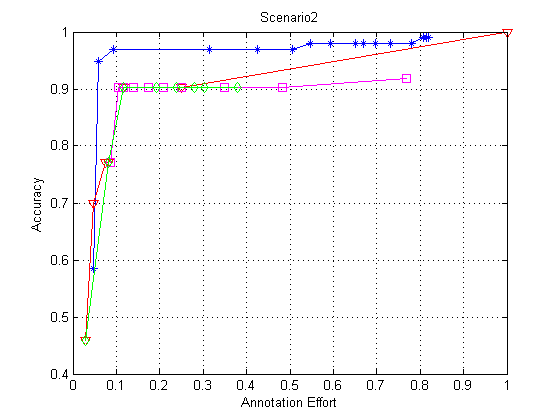}
                \caption{\tiny SVM Classifier}
                \label{fig:svmscenario2}
        \end{subfigure}
         \begin{subfigure}[b]{0.25\textwidth}
                \centering
                \includegraphics[width=\textwidth]{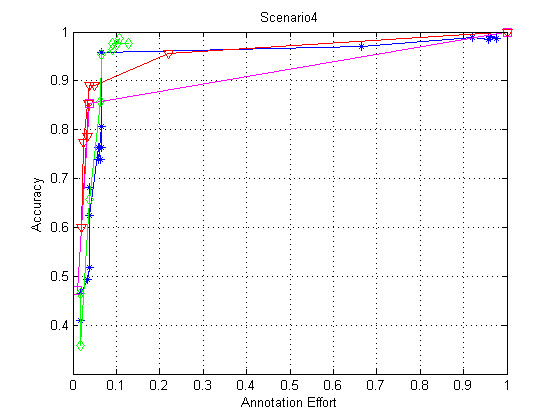}
                \caption{\tiny Naive Classifier}
                \label{fig:naive-scenario4}
        \end{subfigure}%
        ~  
        \begin{subfigure}[b]{0.25\textwidth}
                \centering
                \includegraphics[width=\textwidth]{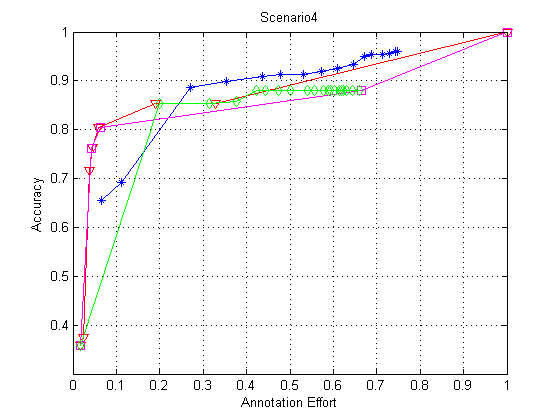}
                \caption{\tiny SVM Classifier}
                \label{fig:naive-scenario4}
        \end{subfigure}
        ~  
        \begin{subfigure}[b]{0.35\textwidth}
                \centering
                \includegraphics[width=\textwidth]{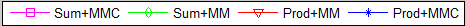}
                                \label{fig:legendmultiplesetting}
        \end{subfigure}
        
                \caption{\small Multiple configurations tested on the synthetic scenarios. ``SUM", ``Prod", ``MMC", and ``MM" indicate sum rule, product rule, modified most confident and modified margin.}\label{fig:setting of synthetic scenario.}
\end{figure}

Figure~\ref{fig:setting of synthetic scenario.} presents the results of multiple settings on Scenarios~\MakeUppercase{\romannumeral 1},...,~\MakeUppercase{\romannumeral 4}.
 One prominent observation on all these results is that using geometric mean (Prod) to combine information of frames in a given batch and the modified most measure (MMC) to select most informative batches give the best performance. 

\begin{figure}
        \centering
        \begin{subfigure}[b]{0.25\textwidth}
                \centering
                \includegraphics[width=\textwidth]{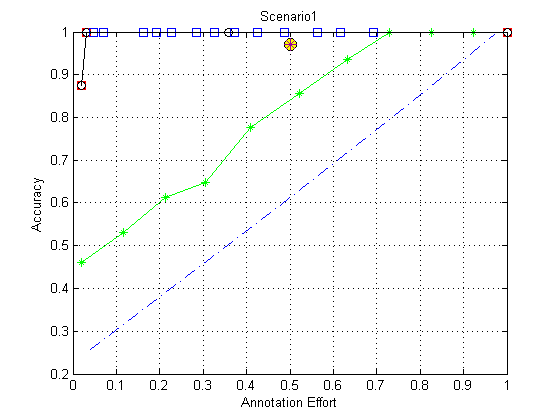}
                \caption{\tiny Scenario\MakeUppercase{\romannumeral 1}}
                \label{fig:scenario1}
        \end{subfigure}%
        ~  
        \begin{subfigure}[b]{0.25\textwidth}
                \centering
                \includegraphics[width=\textwidth]{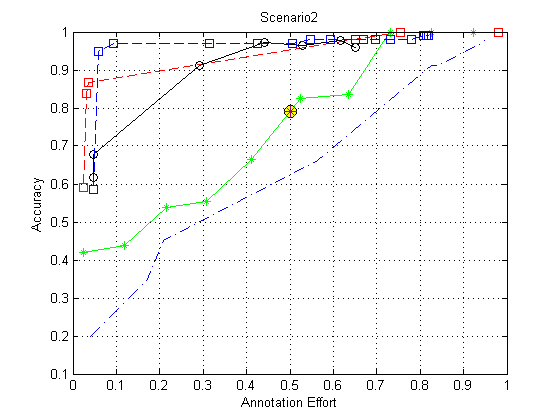}
                \caption{\tiny Scenario\MakeUppercase{\romannumeral 2}}
                \label{fig:scenario2}
        \end{subfigure}
         \begin{subfigure}[b]{0.25\textwidth}
                \centering
                \includegraphics[width=\textwidth]{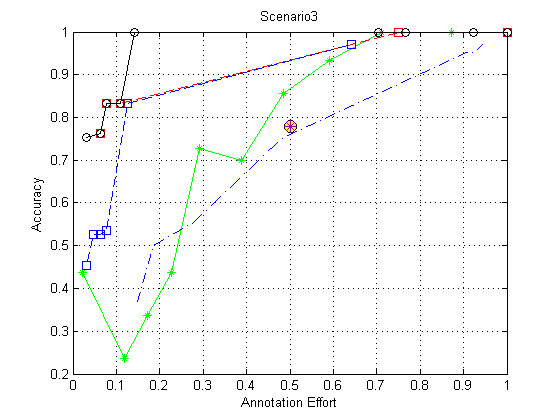}
                \caption{\tiny Scenario\MakeUppercase{\romannumeral 3}}
                \label{fig:scenario4}
        \end{subfigure}%
        ~  
        \begin{subfigure}[b]{0.25\textwidth}
                \centering
                \includegraphics[width=\textwidth]{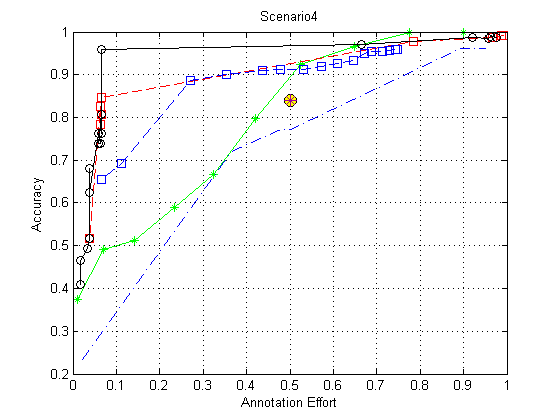}
                \caption{\tiny Scenario\MakeUppercase{\romannumeral 4}}
                \label{fig:scenario4}
        \end{subfigure}
        ~  
        \begin{subfigure}[b]{0.5\textwidth}
                \centering
                \includegraphics[width=\textwidth]{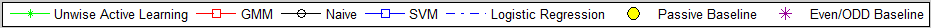}
                                \label{fig:legendmultiplemethod}
        \end{subfigure}
        
                \caption{\small Multi-class classifier comparison on synthetic scenarios using the best configuration (Prod+MM).}\label{fig:multiple classifier synthetic scenario}
\end{figure}

Figure~\ref{fig:multiple classifier synthetic scenario} illustrates the comparative results across multiple classifiers on \emph{Scenarios} \MakeUppercase{\romannumeral 1},..., \MakeUppercase{\romannumeral 4} from which we can observe that: a) NEVIL achieves more than 90\% accuracy with less than 15\% annotation effort in all the datasets, which obviously outperforms baseline approaches. For all the sets, we reached equal accuracy to \emph{Passive} as well as\emph{Even/Odd} Learning while spending much less human resources. b) Naive classifier gives the best overall performance which emphasise the more flexible nature of generative models than discriminative ones. Needless to say, following the results depicted in Figure~\ref{fig:setting of synthetic scenario.} we only present the result of winner setting.

\begin{figure}
        \centering
        \begin{subfigure}[b]{0.25\textwidth}
                \centering
                \includegraphics[width=\textwidth]{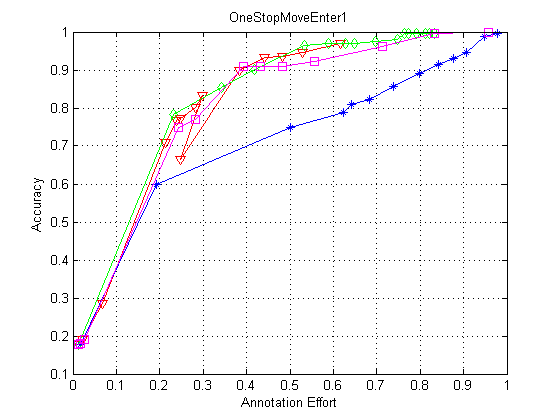}
                \caption{\tiny Logistic Regression Classifier}
                \label{fig:regressiononvideo1}
        \end{subfigure}%
        ~  
        \begin{subfigure}[b]{0.25\textwidth}
                \centering
                \includegraphics[width=\textwidth]{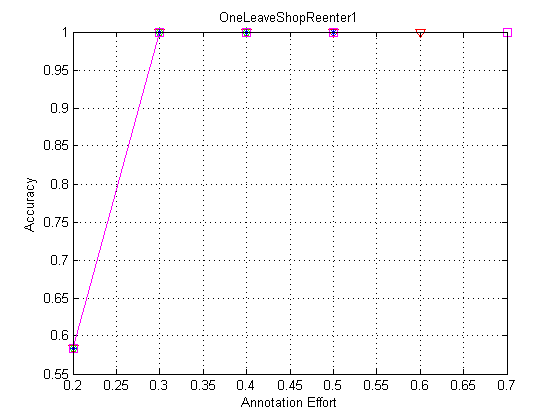}
                \caption{\tiny Logistic Regression Classifier}
                \label{fig:regressiononvideo2}
        \end{subfigure}
         \begin{subfigure}[b]{0.25\textwidth}
                \centering
                \includegraphics[width=\textwidth]{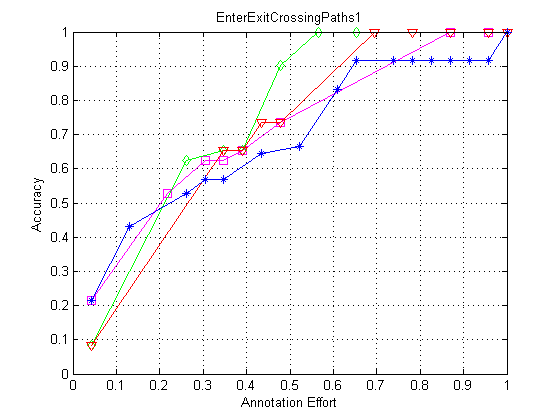}
                \caption{\tiny SVM Classifier}
                \label{fig:svmonvideo1}
        \end{subfigure}%
        ~  
        \begin{subfigure}[b]{0.25\textwidth}
                \centering
                \includegraphics[width=\textwidth]{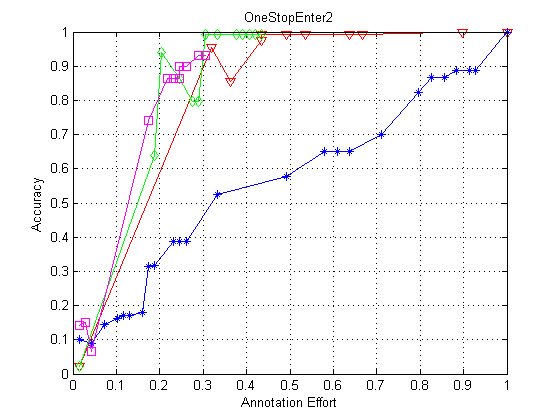}
                \caption{\tiny SVM Classifier}
                \label{fig:svmonvideo2}
        \end{subfigure}
        ~
         \begin{subfigure}[b]{0.35\textwidth}
                \centering
                \includegraphics[width=\textwidth]{legendmultiplesetting.png}
                                \label{fig:legendmultiplesetting}
        \end{subfigure}
        
                \caption{\small Multiple configurations tested on the CAVIAR sequences.``SUM", ``Prod", ``MMC", and ``MM" indicate sum rule, product rule, modified most confident and modified margin.}\label{fig:setting of realdata}
\end{figure}

Figure~\ref{fig:setting of realdata} shows the comparative results on some CAVIAR sequences with various NEVIL configurations. We observe that unlike synthetic scenarios, employing arithmetic mean (SUM) as combination method and modified margin (MM) as selection criteria present winner results. The presence of challenging noise in real data explains the different behaviour of the framework.

\begin{figure*}
        \centering
                \begin{subfigure}[b]{0.3\textwidth}
                \centering
                \includegraphics[width=\textwidth]{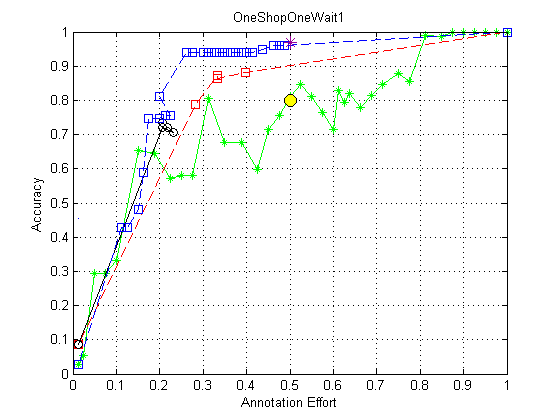}
                \caption{\tiny OneShopOneWait1}
                \label{fig:OneShopOneWait1}
        \end{subfigure}
        ~
         \begin{subfigure}[b]{0.3\textwidth}
                \centering
                \includegraphics[width=\textwidth]{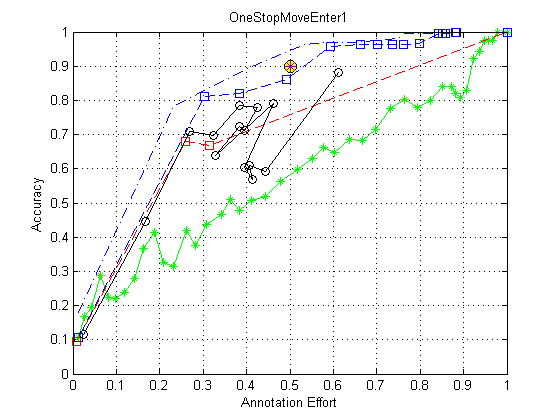}
                \caption{\tiny OneStopMoveEnter1}
                \label{fig:OneStopMoveEnter1}
        \end{subfigure}%
        ~  
        \begin{subfigure}[b]{0.3\textwidth}
                \centering
                \includegraphics[width=\textwidth]{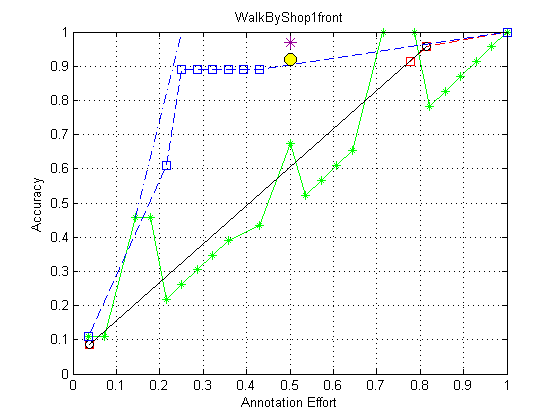}
                \caption{\tiny WalkByShop1front}
                \label{fig:WalkByShop1front}
        \end{subfigure}
        ~  
          
        \begin{subfigure}[b]{0.5\textwidth}
                \centering
                \includegraphics[width=\textwidth]{legendmultiplemethod.png}
                                \label{fig:legendmultiplemethod}
        \end{subfigure}
                        \caption{\small Performance using multiple configurations on the CAVIAR sequences.}\label{fig:multplecaviar}
\end{figure*}

Figure~\ref{fig:multplecaviar} presents the  performance of NEVIL employing various classifiers on multiple CAVIAR sequences. The NEVIL framework achieves over 80\% accuracy with less than 25\% of labelling and in most cases, that is clearly superior to baseline methods. Contrary to results obtained from synthetic data, Discriminative models outperforms than Generative ones. Higher dimension of video streams (herein, equal to 85) may explain this behaviour. Generative models are commonly trained using Maximum-Likelihood Estimation (MLE) that especially for high dimensional data, the likelihood can have many local maxima. Thus, finding the global maximum affects the performance and renders the approach less practical.

\begin{figure}
\begin{center}
\centerline{\includegraphics[scale=0.6]{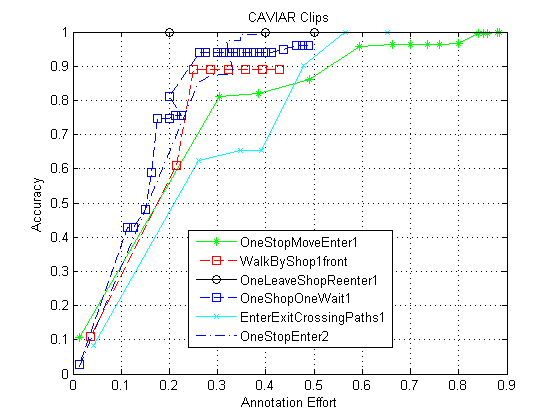}}
\caption{\small Performance of NEVIL on multiple CAVIAR sequences. The results were obtained with the SUM+MM configuration and applying SVM as the classifier.}
\label{fig:svmallcaviarclips}
\end{center}
\end{figure}

Finally, Figure~\ref{fig:svmallcaviarclips} presents the results obtained across multiple CAVIAR scenarios from the most successful setting, which means SVM, SUM, and MMC as base classifier, combination rule and selection criteria, respectively. Under such setting, NEVIL achieves 80\% accuracy with 30\% annotation effort for \emph{OneStopMoveEnter1}, the most complex scenario with 42 streams from 14 classes.

\section{Conclusions} \label{Conclusions}
In this paper, we address the problem of learning from visual streams generated in a multi-camera surveillance scenario. We look at the problem as mining parallel high-dimensional data.
Inspired from active learning strategies, in our proposed framework (NEVIL) an oracle provides labelled batches; multiple informativeness measures are used to determine when the oracle is used.
As base learners are bottlenecks of any learning pipeline, various groups of classifiers were studied and experimentally evaluated. We ran the experiments on synthetic as well as real datasets.

 In synthetic scenarios, where low dimensional clean data is available, applying the geometric mean and the modified most confident measure gives the best and least expensive (in terms of annotation cost) results. However, to get the highest accuracy from noisy visual data we need to apply arithmetic mean for combining information and modified margin to select the most informative batches.

Another question we tried to answer was which classifier to use on a given dataset. In a low dimensional clean data, generative approaches give the best results, however obtaining robust and stable results from high dimensional data is too difficult, as shown by our experiments. The best performance is obtained through discriminative approaches.

While empirical results demonstrated the functionality of the framework, we are currently working the controlling the complexity of the framework which makes it applicable for never-ending scenarios. We would also like to employ special features of video streams generated in a surveillance scenario in order reduce queries as many as possible.

\begin{acknowledgements}
The authors would like to thank Funda\c{c}\~{a}o para a Ci\^{e}ncia e a Tecnologia (FCT)-Portugal for financing this work through the grant SFRH/BD/80013/2011.
\end{acknowledgements}

\bibliographystyle{spbasic}      
\bibliography{example_paper}

\end{document}